# EXPLORING VARIATIONAL AUTOENCODERS FOR MEDICAL IMAGE GENERATION: A COMPREHENSIVE STUDY

[1]KHADIJA RAIS, [2]MOHAMED AMROUNE, [3]ABDELMADJID BENMACHICHE, [4]MOHAMED YASSINE HAOUAM

[1,2,4]Laboratory of mathematics, informatics and systems (LAMIS), Echahid Cheikh Larbi Tebessi University, Tebessa, 12002, Algeria.
[3]Department of Computer Science, LIMA Laboratory, Chadli Bendjedid, University, El-Tarf, PB 73, 36000, Algeria.
Email: [1]khadija.rais@univ-tebessa.dz; [2]mohamed.amroune@univ-tebessa.dz; [3]benmachiche-abdelmadjid@univ-eltarf.dz, [4]mohamed-yassine.haouam@univ-tebessa.dz.

**Abstract** - Variational autoencoder (VAE) is one of the most common techniques in the field of medical image generation, where this architecture has shown advanced researchers in recent years and has developed into various architectures. VAE has advantages including improving datasets by adding samples in smaller datasets and in datasets with imbalanced classes, and this is how data augmentation works. This paper provides a comprehensive review of studies on VAE in medical imaging, with a special focus on their ability to create synthetic images close to real data so that they can be used for data augmentation. This study reviews important architectures and methods used to develop VAEs for medical images and provides a comparison with other generative models such as GANs on issues such as image quality, and low diversity of generated samples. We discuss recent developments and applications in several medical fields highlighting the ability of VAEs to improve segmentation and classification accuracy.

## I. INTRODUCTION

Medical image generation is essential to the advancement of medical applications, particularly in the areas of diagnosis, treatment planning, and clinical trials. However, the lack of annotated data is a major obstacle to the advancement of many medical imaging applications, mainly due to privacy laws and the costs associated with manual annotation, as well as the limited availability of public datasets. The lack of data reduces the performance of machine learning models, particularly deep learning approaches that require large, comprehensive datasets to perform optimally.

VAEs, a type of generative model that has become increasingly popular over the last years, have proven to be an architecture capable enough to overcome this challenge. VAEs can perform both tasks: disentanglement and new sample generation, making them a strong candidate for use in generating synthetic data specific to medical images. Such synthetic datasets will help to create better training sets, promote model generalization, and overcome overfitting in data-scarce domains.

VAE could improve AI analysis tasks including classification and segmentation. By combining HVAE and discriminative regularization, Kebaili et al were able to create realistic images and their high-quality masks using both BRATS MRI and HECKTOR PET datasets to improve segmentation[1]. Using VAE, Elbattah et al. generated synthetic images based on eye-tracking results, which demonstrated that VAE improves classification accuracy[2].

In this paper, we focus on the application of VAEs to medical image generation and conduct a comprehensive study on their ability to generate realistic and diverse images. We review the details of the typical architecture and evaluation metrics. In this regard, we focus on VAE-based strategies to highlight their potential to generate near-photorealistic synthetic medical images and thus address one of the most critical issues: the urgent need for data scarcity in medical imaging.

## II. VARIATIONAL AUTOENCODER

In 2013, Kingma and Welling developed the Variational Autoencoder (VAE), they integrate probabilistic modeling into the foundations of deep learning. The principle of VAE is to generate new data from input data via a probabilistic latent space[3].

**The main elements of VAE are[4]:**
Encoder:Using input data x, the encoder extracts latent variables and produces a vector representing latent space z.

Latent space:z is the latent space that contains the encoded data and serves as the output of the encoderand the input of the decoder.

Decoder:x' is created from the output z, where x' is the generated data.

Note that the encoder and decoder are neural networks, and the latent space is a distribution of variables.

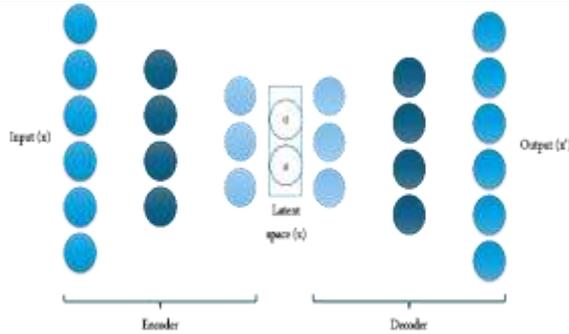

**Figure 1: The overview of VAE**

VAEs generate diverse samples by encoding inputs into a latent space, helping to reduce overfitting and improve model robustness. They suffer less from mode collapse than GANs but may produce less realistic outputs. Despite this, VAEs offer a flexible way to enhance data diversity, even with small datasets[5].

VAE compared to other generated techniques considers that it works better with small amounts of data[6]. Complex data distributions can be handled by VAE by learning a continuous and structured latent space[7].

**Variational autoencoder architecture:**
The researchers have developed several variants of the VAE architecture[8]:
- VAE: Variational autoencoder is the vanilla architecture which introduces three main elements including an encoder which learns the input data, the latent space and a decoder which reconstructs the input data.
- CVAE: Conditional VAE rebuilds the missing areas of the image, and it can also create an image from a class label. In addition, the authors propose a multi-scale output prediction to be used with large-scale images.
- VQVAE: Vector quantized VAE improves the generated image when the decoder input is an embedding vector.
- VAE-GAN: This method introduces VAE within the framework of GAN, by adding a discriminator, and the decoder works with the generator.

EndoVAE is an endoscopic variational autoencoder, a novel technique presented by Diamantis et al. to improve the ability of VAE to generate real medical images and address limited data availability[9].

Based on the Hamiltonian variational autoencoder, the researchers presented an end-to-end architecture to generate medical images and their corresponding segmentation masks. This approach is validated in two datasets (BRATS and HECKTOR)[6]. Lee et al. introduce a VAE-based generative active learning framework to augment radiological data for veterinary medicine, aiming to address data shortage issues in computer-aided diagnosis (CAD) systems[10].

To improve the generation of medical datasets and their corresponding masks, Kebaili et al. combined HVAE with discriminative regularization, this approach efficiently created realistic images and their high-quality masks using both the BRATS MRI image dataset and the HECKTOR PET image dataset[1].

VAE-GAN is a combination of GAN and VAE with the aim of augmenting medical image analysis tasks, showing efficient results and demonstrating high-quality and diversified images[7].

Elbattah et al. used VAE to generate a synthetic image-based representation of the eye-tracking. The results demonstrate the effectiveness of VAE in improving classification[2].

The researchers presented an efficient self-learning mechanism where the LSTM achieves high accuracy by adding images from the original images with the VAE[11].
Rais et al present Disc-GAN[12] an approach that places the VAE in the GAN framework, where the VAE is considered as a generator and the discriminator distinguishes fake images generated with the VAE from real ones, the authors use the vanilla VAE also in another work to improve classification[13].

Ahmad et al, present a hybrid approach containing VAE with GAN to generate brain tumor images, the architecture consists of an encoder-decoder, a generator and a discriminator[14].

Huo et al. propose a progressive Brain Lesion Synthesis Framework PAVAE or progressive variational adversarial autoencoder, they present a 3D progressive variational autoencoder to approximate both shape and intensity information[15].

Kebaili et al cited more VAE architectures including ICVAE, Intro VAE, RH VAE, AL VAE, etc[16].

## III. EVALUATION OF VARIATIONAL AUTOENCODER EFFECTIVENESS

EndoVAE has been evaluated through the accuracy of classification tasks and it overcomes the associated study and shows the effectiveness in the generated endoscopic images[9].

The HVAE-based approach proposed by Kebaili et al evaluated the DSC obtained by training U-Net and they found an improvement in both BRATS asnd HECKTOR datasets. The quantitative performances of image similarity, including PSNR and SSIM, are also evaluated between the test and generated images[6].

Lee et al. evaluated the generated images using FID by measuring the feature dissimilarity between the generated images and the real images. They also used classification measures including F1 score, precision, accuracy, and recall[10].

Kebaili et al in [1]used two categories to evaluate their work, the first is to evaluate the quality of generated images via PSNR, FID and LPIPS and the second is via visibility of generated images, they also evaluate the quality of generated masks using KLD, JSD and DSC.

The quality of the images generated by VAE-GAN is evaluated by FID and IS[7].Elbattah et al. implemented a CNN model for classification and analyzed the accuracy based on the ROC curve[2].The authors in [11] evaluate the model against various benchmarks including sensitivity, specificity, accuracy, and F1-Sc for LSTM classification.

Researchers who create a hybrid architecture between GAN and VAE evaluate the generated images via a classifier and their classification metrics[14].Other researchers[17] also combine the advantages of both GAN and VAE approaches, representing the VAE-GAN model where the generator uses a random noise vector, the discriminator distinguishes real from fake images, the VAE encoder maps the input image to a latent representation, and the decoder uses it to reconstruct the image.

Huo et al. evaluate the generated lesions with SNR, SSIM, and NMSE. In addition, they evaluate the segmentation task after augmentation with four metrics, including Dice coefficient, Jaccard index, ASD, and Hausdorff distance[15].

## IV. DISCUSSION

VAE-based techniques have been shown to generate medical images for classification and segmentation tasks, demonstrating the versatility of VAE in solving data scarcity issues in the healthcare domain. It has been shown that VAE can greatly contribute to overcoming data scarcity in medical image generation, which means gains for classification and also segmentation. Diamantis et al.[9]and Lee et al.[10]have demonstrated synthetic data generation and improved diagnostic accuracy in endoscopy as well as veterinary medical fields using VAE. Similarly, hybrid approaches such as the VAE-GAN model[14]improve brain tumor classification by providing synthetic and realistic data in terms of diversity.

In segmentation, models such as HVAE[1]and PAVAE[15], similarly use VAE for both image generation from scratch under the semi-supervised or unsupervised setting maximizing data-efficiency to generate improved medical images and masks while reducing annotations required for state of art accuracy on many public datasets. These models demonstrate that VAEs are capable of facilitating synthesis high-fidelity samples for training even on large datasets with missing annotations.

While the images generated by a VAE may not be as realistic all the time compared to those from GANs, this feature of reducing overfitting and enriching data diversity is invaluable especially in medical applications when combined with other generative methods.

|  | (Author, year) | techniques | Objective |
|---|---|---|---|
| Classification | (Diamantis et al., 2022) [9] | EndoVAE | Evaluate VAEs as an alternative to GANs for generating synthetic endoscopic images to address limited medical imaging data availability. |
| Classification | (Lee et al, 2023) [10] | VAE-based generative active learning framework | Enhance the performance of CAD systems in veterinary medicine by generating and augmenting radiology data, for improved diagnostic accuracy. |
| Classification | (Elbattah et al, 2021)[2] | VAE | Address data scarcity in healthcare by employing VAEs to generate synthetic eye-tracking data, thereby enhancing performance in classification tasks. |
| Classification | (Naga et al, 2023) [11] | VAE | Leveraging self-learning from both autogenerated and original images to improve diagnostic accuracy. |
| Classification | (Ahmad et al., 2022) | Hybrid approach between | Augmentlimited datasets to improve |
| | [14] | GAN and VAE | the accuracy and effectiveness of brain tumor classification models, |
| Segmentation | (Kebaili et al, 2023)[6] | Novel architecture based on HVAE | Generate diverse and realistic medical images and their corresponding masks by utilizing an HVAE to enhance image generation quality. |
| | (Kebaili et al, 2024) [1] | HVAE with discriminative regularization | Improve medical image segmentation performance by generating diverse and realistic image-mask pairs using the proposed HVAE-based framework. |
| | (Huo et al, 2022) [15] | PAVAE | Address the challenge of limited annotated data for emerging treatments by generating synthetic brain lesion images to enhance segmentation models. |

**Table 1: Summary of VAE-based Techniques for Medical Image Generation in Classification and Segmentation.**

## V. CONCLUSION

In this paper, we propose variational autoencoders as a general tool to address medical image generation from limited annotated data.

Variational autoencoder is particularly effective for generating many different synthetic images, which are needed to augment deep learning-based classification and segmentation models. The high-quality samples that VAEs can generate (by learning meaningful latent representations) in turn reduce overfitting and make the model more stable against adversarial attacks.

Their ability to integrate seamlessly with other models, such as GANs, makes VAE a natural choice for hybrid architectures that use the best of both worlds. While VAE results are not as realistic on average as GANs, they can provide useful data augmentation for small datasets.

Overall, the promising results obtained from various studies underscore the importance of VAEs in medical imaging research. By addressing data scarcity and enhancing the quality of training datasets.

★ ★ ★